\documentclass[a4paper,conference]{IEEEtran}
\usepackage{amsmath,amssymb} 
\usepackage[misc]{ifsym}
\usepackage{graphicx}
\usepackage{cite}
\usepackage{comment}
\usepackage{color}
\usepackage{multirow}
\usepackage{multicol}
\usepackage{booktabs}
\usepackage{subfigure}
\usepackage{epsfig}
\usepackage{makecell}
\usepackage{soul}

\newcommand{\etc}{\textit{etc}.}
\newcommand{\etal}{\textit{et al}. }

\newcommand{\eg}{\textit{e}.\textit{g}.}
\newcommand{\tabincell}[2]{\begin{tabular}{@{}#1@{}}#2\end{tabular}}

\begin{document}
\title{OVC-Net: Object-Oriented Video Captioning with Temporal Graph and Detail Enhancement}

\author{
\IEEEauthorblockN{Fangyi Zhu}
\IEEEauthorblockA{School of Artificial Intelligence\\
Beijing Univerisity of Posts\\and Telecommunications\\
Beijing, China\\
Email: zhufangyi@bupt.edu.cn}
\and
\IEEEauthorblockN{Jenq-Neng Hwang}
\IEEEauthorblockA{Department of Electrical and\\Computer Engineering\\
Univerisity of Washington\\
Seatlle, USA\\
Email: hwang@uw.edu}
\and
\IEEEauthorblockN{Zhanyu Ma} 
\IEEEauthorblockA{School of Artificial Intelligence\\
Beijing Univerisity of Posts\\and Telecommunications\\
Beijing, China\\
Email: mazhanyu@bupt.edu.cn}
\and
\IEEEauthorblockN{Guang Chen}
\IEEEauthorblockA{School of Artificial Intelligence\\
Beijing Univerisity of Posts and Telecommunications\\
Beijing, China\\
Email: chenguang@bupt.edu.cn}
\and
\IEEEauthorblockN{Jun Guo}
\IEEEauthorblockA{School of Artificial Intelligence\\
Beijing Univerisity of Posts and Telecommunications\\
Beijing, China\\
Email: guojun@bupt.edu.cn}
}

\maketitle

\begin{abstract}
Traditional video captioning requests a holistic description of the video, yet the detailed descriptions of the specific objects may not be available. Without associating the moving trajectories, these image-based data-driven methods cannot understand the activities from the spatio-temporal transitions in the inter-object visual features. Besides, adopting ambiguous clip-sentence pairs in training, it goes against learning the multi-modal functional mappings owing to the one-to-many nature. In this paper, we propose a novel task to understand the videos in object-level, named object-oriented video captioning. We introduce the video-based object-oriented video captioning network (OVC)-Net via temporal graph and detail enhancement to effectively analyze the activities along time and stably capture the vision-language connections under small-sample condition. The temporal graph provides useful supplement over previous image-based approaches, allowing to reason the activities from the temporal evolution of visual features and the dynamic movement of spatial locations. The detail enhancement helps to capture the discriminative features among different objects, with which the subsequent captioning module can yield more informative and precise descriptions. Thereafter, we construct a new dataset, providing consistent object-sentence pairs, to facilitate effective cross-modal learning. To demonstrate the effectiveness, we conduct experiments on the new dataset and compare it with the state-of-the-art video captioning methods. From the experimental results, the OVC-Net exhibits the ability of precisely describing the concurrent objects, and achieves the state-of-the-art performance.
\end{abstract}

\IEEEpeerreviewmaketitle

\section{Introduction}

Video captioning, which bridges two modalities: vision and language, poses an intriguing challenge for artificial intelligence \cite{1,2,3,4,5,6,7}. Besides, video captioning has significance to human-robot interaction and visually impaired people \cite{6,7}.

\begin{figure}[t]
\centering
\includegraphics[width=\linewidth]{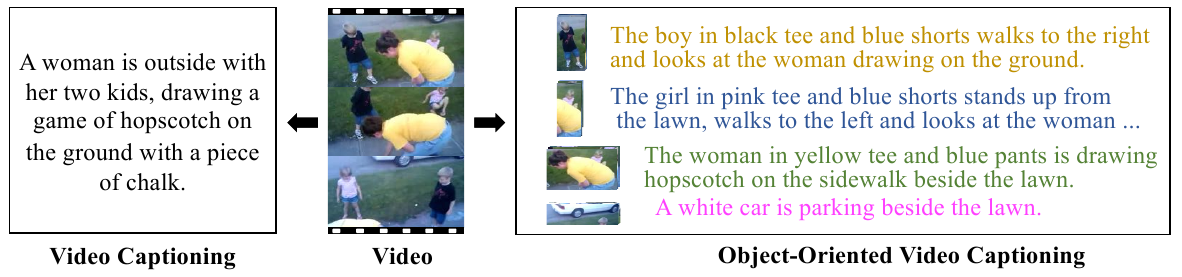}
\caption{In natural videos, there are multiple concurrent objects and activities. We propose the object-oriented video captioning to understand the video in object-level, describing concurrent objects and activities in more details.}
\label{fig:task_compare}
\end{figure}

Traditional video captioning calls for a holistic description of the entire video or a random object, unable to process concurrent activities. Actually, while human watching video, we focus on different objects based on personal interests. Thus, comparing to giving a monotonous and inconclusive description, it seems to be more meaningful to allow detailed descriptions for multiple objects.

So far, most methods of video captioning are frame-level based, where the objects in consecutive frames are independent without temporal associations \cite{1,2,3,4,5,6,7,11,13,15,16,17,18,20,46}. Therefore, besides the visual features, most works adopt the models for action recognition \cite{31,32,33} to extract the spatio-temporal 3D convolutional (C3D) features as temporal cues \cite{11,15,16,18,20,46}. However, using a pre-trained model for action recognition to extract temporal information on our scenario brings two problems: one is it cannot extract effective features under multi-objects due to in most videos for action recognition there only exist a single object; the other is, for the model can only adapt to human actions, it cannot capture the temporal information of the non-human objects, \eg, animals, vehicles. Moreover, the C3D features and the visual features, fed into the captioning module at the same time, usually represent different periods, resulting in confusion during training.

Besides, all datasets for video captioning provide video-sentence pairs without information associated with objects, where each video has multiple sentences for different objects \cite{18,19,20,27}. The indefinite one-to-many video-sentence pairs are difficult for the captioning system to bridge the vision contents and visual words. Hence, especially when multiple objects exist, previous methods fail to distinguish the concurrent objects. Furthermore, these methods depend on training under large amounts of data. Whereas, in many application scenarios, there are not available sufficient data, resulting in limited applicability of these data-driven methods.

For addressing the problems above, we first propose a novel task, named object-oriented video captioning, transforming the video-level captioning to object-level. Instead of a rough description of the entire video, we aim at understanding the video in object-level, which is closer to human thinking while watching videos. Understanding the activities in object-level leads to a greater understanding of the videos, deserving more of our attentions. Then, to kick-off the research in this novel area and facilitate learning the vision-language functional and translational relationships, we re-annotate a new database, providing consistent object-sentence pairs associated with identities. Finally, we design the video-based object-oriented video captioning network (OVC)-Net via temporal graph and detail enhancement. Its main novelties and advantages are:

{\bf Object-oriented temporal graph.} The object-oriented temporal graph is built to understand the spatio-temporal evolution in the trajectories to wean off the dependence on training under large amounts of data. It achieves the motivation of reasoning activities along time without any supplementary cues from other tasks, allowing the proposed approach turn into real video understanding.

{\bf Detail enhancement.} The construction of detail enhancement module effectively facilitates learning the discriminative features among different objects, thereby the generated descriptions can be more accurate and detailed.

We conduct experiments on the new dataset and compare with the state-of-the-arts of video captioning. Experimental results demonstrate that the proposed approach can achieve the state-of-the-art performance in terms of BLEU@4, METEOR, CIDEr-D and ROUGE-L metrics \cite{41,42,43,44}. More importantly, the novel task can analyze the video in object-level rather than giving a holistic description of a single activity.


\section{Related Works}

Video captioning requests a systematic description of the videos. When image captioning gradually getting matured \cite{8,9,10,14}, more attentions shift to video captioning \cite{11,12,13,15,16,17,18}. Recently, a large number of methods have been proposed for video captioning, where the encoder-decoder architecture have been widely adopted \cite{6,7,8,10,11,12,13,15,16,17,18,36,37,38,39,60,61,62}.

With the development of attention mechanism, Xu \etal and Liu \etal designed a spatial attention to automatically exploit the impact of different regions in each frame \cite{6,8,10}. Meanwhile, a lot of works explored the temporal attention to select the key frames for the current word generation \cite{15,19,38}. Considering most works dependence on the forward flow (video-to-sentence), Wang \etal referred to the idea of dual learning and proposed RecNet \cite{7} to exploit the backward flow (sentence-to-video). On top of captioning system, RecNet stacked another module to reconstruct the visual contents.

However, most methods still execute as image-based and utilize the inter-tangled frame-level features of all objects and stuff \cite{1,2,3,4,5,6,7,11,13,15,16,17,18,19,20,28}. The frame-level features for each frame are independent without definite association. To explore the moving trajectories, some works introduced tracking operations into video captioning. Zhang \etal proposed the object-aware aggregation with a bidirectional temporal graph (OA-BTG) \cite{46} to track the salient objects in the video. However, the tracking operation is not precise enough since it associates the objects among few pre-selected objects in each frame. In addition, it performed tracking on sampled frames rather than the original video, which may also yield mistakes. More crucially, it merged all the objects by mixing their visual features without associating the trajectories when fed to the subsequent captioning module, which is still difficult for understanding the spatio-temporal transition.

Recently, to describe concurrent events, Li \etal \cite{18} proposed dense video captioning, which bridges two separate tasks: temporal action location and video captioning. Dense video captioning requires to locate a set of clips where events happen and describe the predicted clips. Despite it can process concurrent activities, it still cannot analyze specific objects.

Overall, previously proposed video captioning methods fail to provide precise and sufficient information for object-level analyses. Despite significant improvements have been achieved, most approaches are data-driven and image-based, incapable of reasoning the step-by-step object-level activities along the time domain.

\section{The Proposed OVC-Net}

\begin{figure*}[!t]
\centering
\includegraphics[width=\linewidth]{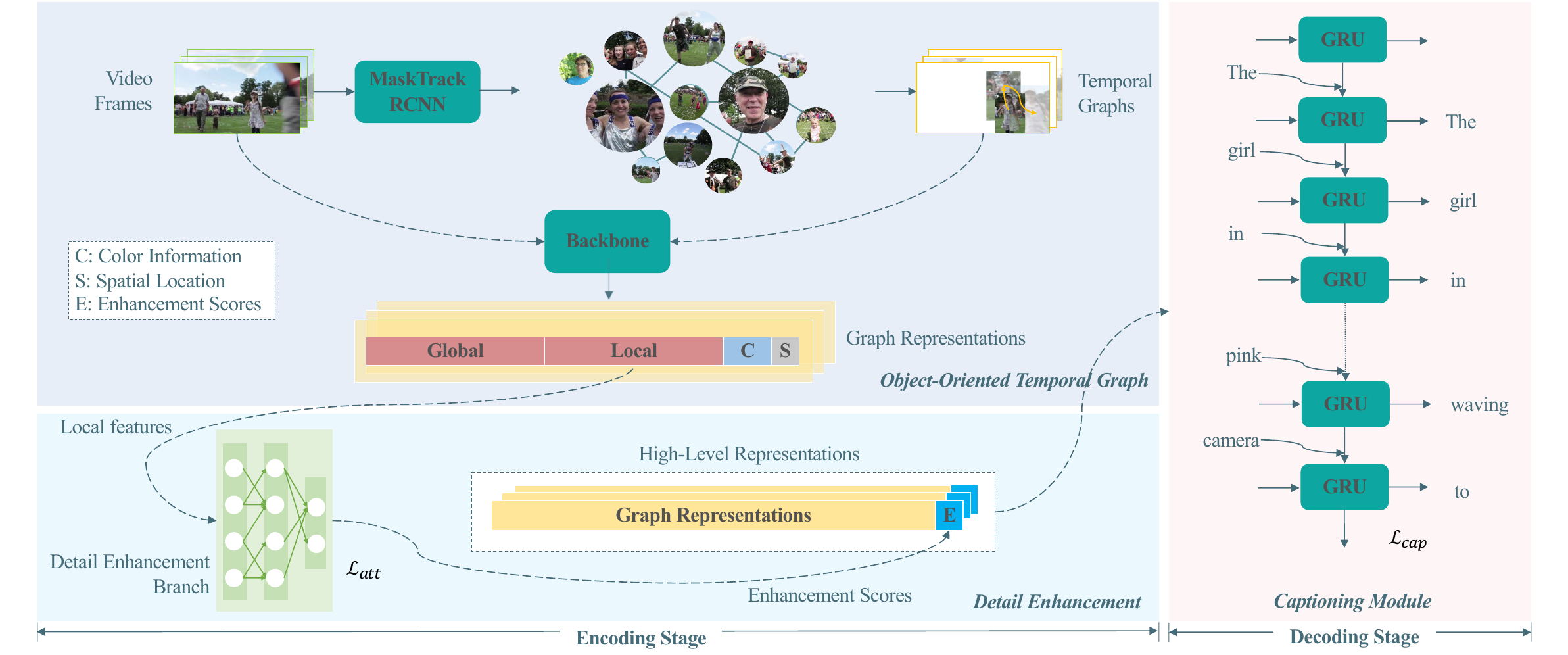}
\caption{Overview of the proposed OVC-Net. The OVC-Net consists of three modules: first build the object-oriented temporal graphs, second feed the local features into the detail enhancement module to capture the discriminative features for different objects and generate the final high-level representations, third yield the captions word-by-word with the obtained high-level representations in the captioning module.}
\label{fig:flowchart}
\end{figure*}

In this section we introduce our main framework for object-oriented video captioning. As illustrated in Fig. \ref{fig:flowchart}, the OVC-Net is based on the encoder-decoder structure and consists of three modules: {\bf (1) Object-oriented temporal graph:}  First conduct tracking to obtain a set of object trajectories, then construct temporal graphs to represent each object. {\bf (2) Detail Enhancement:} Feed the local information from module (1) into the detail enhancement branch to capture more discriminative features among different objects. The obtained enhancement scores are combined with the temporal graphs from module (1) as high-level representations of objects. {\bf (3) Captioning module:} Decode the high-level representations from module (2) into sentences word-by-word. In the following subsections, we will address the three modules in turn.

\subsection{Object-Oriented Temporal Graph}

In this subsection, we introduce how to build the object-oriented temporal graph. Briefly, this module consists of three steps: first, conduct detection and tracking to get the object trajectories. Second, the local information and global information of the objects at different frames are obtained. Third, the temporal graph for each object is built. In the following, we will present the three steps successively.

Recently, many excellent works for object-level visual analyses have emerged, \eg , Faster R-CNN, YOLO, Mask R-CNN for object detection and Tracktor, TrackletNet Tracker(TNT), the MaskTrackRCNN for multi-object tracking \cite{21,22,25,47,24,48}. The MaskTrackRCNN can conduct object detection, instance segmentation and object tracking at the same time. Moreover, it is trained on natural videos, which are similar to our scenarios, while most tracking methods are trained on scenes for video surveillance or autonomous driving. Thus, we adopt MaskTrackRCNN as a combined detector-tracker in this work.

First, given a $T$-frame video $V=\left \{ v^{1},\dots,v^{T} \right \}$, we adopt MaskTrackRCNN to obtain the trajectories for all objects as
\begin{align}
    \label{eq:build_graph_!} O&=F_{MTRCNN}(V),\\
    \label{eq:build_graph_2} o&=\left \{ v_{o},b_{o} \right \},
\end{align}
where $F_{MTRCNN}$ is the MaskTrackRCNN model. $O=\left\{ o_{1},\dots,o_{n}\right \}$ is a set of detected objects. For each object $o$, we record the time stamps $v_{o}$ when it occurs and the corresponding spatial locations where it exists $b_{o}$. $v_{o}=\left \{ v^{t_{1}},\dots,v^{t_{j}},\dots,v^{t_{m}} \right \}$ denotes a set of frames where the object $o$ occurs, $m$ is the total number of the frames where the object $o$ exists. $b_{o}=\left \{ b_{o}^{t_{1}},\dots,b_{o}^{t_{j}},\dots,b_{o}^{t_{m}} \right \}$, where $b_{o}^{t}=\left [  x_{o}^{t},y_{o}^{t},w_{o}^{t},h_{o}^{t}\right ]$ is the spatial location of $o$ at time $t$.

\begin{figure}[t]
    \centering
    \includegraphics[width=\linewidth]{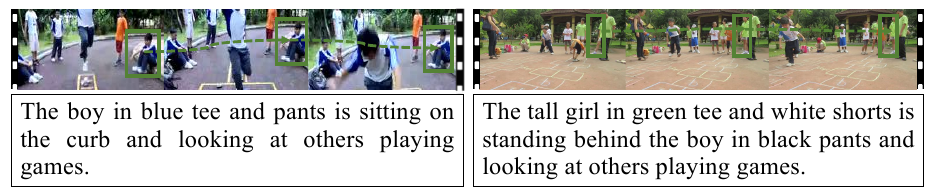}
    \caption{Examples of the paired local and global information.}
    \label{fig:global_local_pair}
\end{figure}

Second, for each object, we extract its local and global features as
\begin{align}
   \label{eq:build_graph_3} \phi_{lo}^{t}=\left[\Theta\left(o^{t}\right), c_{o}^{t}\right]&=\left[\Theta\left(v^{t}, b_{o}^{t}\right), c_{o}^{t}\right], \\
    \label{eq:build_graph_4} \phi_{go}^{t}&=\Theta\left(v_{o}^{t}\right),
\end{align}
where $\Theta$ is a pre-trained neural network for extracting the visual semantic features. We feed the cropped object into $\Theta$ to get its local visual features $\Theta\left(o^{t}\right)$. However, the features from the top layers of a neural network lack the detailed appearance information, which is important for distinguishing different individuals, we thus combine the color histograms $c_{o}^{t}$ into the local visual features. Finally, the local features $\phi_{lo}^{t}$ of $o$ at $t$ consist of two components: $\Theta\left(o^{t}\right)$ and $c_{o}^{t}$ as shown in~\eqref{eq:build_graph_3}.

To incorporate the interactions of a tracked object with other objects and the stuff, we again adopt $\Theta$ to extract the global visual features $\phi_{go}^{t}$ where the object occurs in~\eqref{eq:build_graph_4}. Our approach is quite different from other works which directly adopt frame-level features. In previous works, the frame-level information are utilized to learn the actions as well. In our work, with the foreground objects and stuff being separated, the activities are learned by analyzing the temporal transitions in local features along time. The global features serve as a supplement for the object-object interactions and the object-stuff interactions. Some works, \eg, the Fine-Grained Spatial Temporal Attention Model (FSTA) \cite{6}, adopts Mask R-CNN \cite{25} to filter the foreground objects from the stuff, and completely gives up the stuff. In fact, stuff makes up the majority of our visual surroundings, \eg, road, sky, grass, beach. It is useful to infer the positions and orientations of the objects, as well as the interactions with other objects and the stuff, and thus is crucial for scene understanding. Besides, some works adopt the C3D features as local information to learn activities and frame-level features as global supplement \cite{11,15,16,18,20}. However, the global features and the C3D features at the same time step, which are jointly fed into the subsequent captioning module, always represent different periods of the video. Hence, the model cannot effectively discover the relationships between the local and the global. In our work, we use the features of the frame where the object exists as global supplement, which is more effective for the network to learn the inherent relationships between local and global. As shown in Fig. \ref{fig:global_local_pair}, our paired local and global features are  more informative for learning the interactions between different objects or between objects and the stuff.

To better exploit the spatial relationships between local and global, we further combine the spatial locations of the objects $b_{o}^{t}$ in the joint local and global features $\phi_{o}^{t}=\left [  \phi_{lo}^{t},\phi_{go}^{t}\right ]$. Moreover, the explicit spatial movement can significantly facilitate understanding the activities, \eg, jump, walk, squat.

\begin{figure}[t]
\centering
\includegraphics[width=\linewidth]{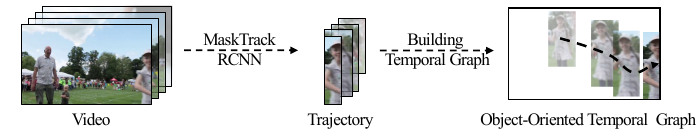}
\caption{An example of building an object-oriented temporal graph for an object. Actually, we build the temporal graph for each object in the video.}
\label{fig:build_temporal_graph}
\end{figure}

Third, with the results obtained in the previous steps, the object-orient temporal graph $G_{o}$ is built to present the object $o$. The graph representation $g_{o}^{t}=\left[\phi_{o}^{t}, b_{o}^{t}\right]$ of $o$ at $t$ consists of the combined feature $\phi_{o}^{t}$ and the spatial location $b_{o}^{t}$. In Fig. \ref{fig:build_temporal_graph}, we show the process of building a temporal graph for an object. Actually, we build the temporal graph for each object in the video, thus we can describe all concurrent objects.


\subsection{Detail Enhancement}

Object-oriented video captioning requests more informative captions to distinguish different objects, incorporating more detailed descriptions, such as gender. We thus design the detail enhancement branch in Fig. 2, with which the whole framework can learn more discriminative features among different objects and identities in the backpropagation.

Given the local features $\phi_{lo}^{t}$ of object $o$ at time $t$, $m$ is the number of frames where the object exists, we have
\begin{equation}
\gamma_{o}=\mathcal{F}_{DE}\left(W_{d}\sum_{t_{1}}^{t_{m}}\phi_{lo}^{t_{j}} / m+b_{d}\right),
\label{eq:classification_1}
\end{equation}
where $\mathcal{F}_{DE}$ is detail enhancement branch, consists of fully connected layers, and $\gamma_{o}$ is the obtained enhancement scores. $W_{d}$ and $b_{d}$ separately denote the learned parameters and bias, respectively. Each dimension of $\gamma_{o}$ represents the probability of $o$ associated with each class. The categorical cross-entropy is adopted as the loss function $\mathcal{L}_{DE}$. Then, the enhancement scores are concatenated with the temporal graph $g_{o}^{t}$ as the high-level presentation $H_{o}^{t}=\left[g_{o}^{t}, \gamma_{o}\right]$, which is the final input of the subsequent captioning module of object $o$ at time $t$.

\subsection{Captioning Module}

Given a high-level presentation $H$ of an object, the captioning module is required to understand its visual contents and automatically yields the description $s=\left\{s_{1},\dots,s_{K}\right\}$ word-by-word in~\eqref{eq:captioning_1}, where $\theta$ denotes the parameters learned, and $K$ is the length of the description. $s_{1},\dots,s_{k-1}$ represent the generated partial words. During training, the parameters $\theta^{*}$ are learned by maximizing~\eqref{eq:captioning_2}.
\begin{align}
\label{eq:captioning_1} \mathrm{P}(\mathrm{S} | \mathrm{H})&=\prod_{k=1}^{K} P\left(s_{k}|s_{1},s_{2},\dots, s_{k-1}, H ; \theta\right). \\
\label{eq:captioning_2} \theta^{*}&=\underset{\theta}{\operatorname{argmax}} \sum_{(H,S)} \log p(S|H;\theta).
\end{align}

A RNN has the capability that generates a sequence, therefore most works adopt LSTMs or GRUs in the captioning module \cite{39,40}. In our scheme, we choose the GRU, as it has fewer parameters and is easier to converge with less data. A GRU has two gates: update gate and reset gate. The reset gate decides how much past information to forget, and the update gate controls what information to throw away and what information to carry over. In brief, the GRU is updated by~\eqref{eq:captioning_gru}, where $H_{k}$ and $z_{k-1}$ denote the current input and the previous hidden states, respectively.
\begin{equation}
z_{k}=\text{GRU}\left(H_{k}, z_{k-1}\right).
\label{eq:captioning_gru}
\end{equation}
In addition, we adopt the temporal attention mechanism to help in deciding which frames are the key frames for the current word generation, and avoiding the negative impact of incorrect tracking results (\eg, ID switch).

Given a target sentence $s^{*}=\left\{s_{1}^{*},\dots, s_{K}^{*}\right\}$, we train the captioning module by minimizing the negative log likelihood in~\eqref{eq:loss_captioning}. The detail enhancement branch and the captioning module are trained jointly. Finally, the loss of our full framework is calculated by~\eqref{eq:loss_total}, where $\lambda$ is a hyper-parameter to balance the detail enhancement branch,
\begin{align}
   \label{eq:loss_captioning} \mathcal{L}_{CAP}(\theta)&=-\sum_{k=1}^{K} \log \left(p_{\theta}\left(s_{k}^{*} | s_{1}^{*}, s_{2}^{*}, \ldots, s_{k-1}^{*}\right)\right),\\
   \label{eq:loss_total} \mathcal{L}&=\mathcal{L}_{CAP}+\lambda\mathcal{L}_{DE}.
\end{align}

\section{Experimental results and discussions}

\begin{table*}[!t]
\caption{Comparisons with the standard datasets for video description. We group all objects which may have activities into three super-classes: human, animal and vehicle. Except FSN, which focuses on human actions only, the other datasets require to describe all the three super-classes. }
\resizebox{\linewidth}{!}{
\renewcommand\arraystretch{1.2}
\begin{tabular}{cccccccccc}
\hline
Dataset & Task & Data Type &
  \begin{tabular}[c]{@{}c@{}}Length of\\ Video (sec)\end{tabular} &
  \begin{tabular}[c]{@{}c@{}}Object\\  Classes\end{tabular} &
  \begin{tabular}[c]{@{}c@{}}Length of\\ Sentence\end{tabular} &
  \begin{tabular}[c]{@{}c@{}}Verbs per \\ Sentence\end{tabular} &
  \begin{tabular}[c]{@{}c@{}}Verbs\\ Ratio (\%)\end{tabular} &
  \begin{tabular}[c]{@{}c@{}}Adjectives per \\ Sentence\end{tabular} &
  \begin{tabular}[c]{@{}c@{}}Adjectives \\ Ratio (\%)\end{tabular} \\ \hline
MSR-VTT    & video captioning           & clip-sen   & 20   & 3 & 9.28  & 1.37 & 14.80 & 0.66 & 24.83 \\
MSVD       & video captioning           & clip-sen   & 10  & 3 & 8.67  & 1.33 & 19.60 & 0.25 & 17.48 \\
ActivityNet & dense video captioning           & clip-sen   & 180 & 3 & 13.48 & 1.41 & 10.40 & 0.67 & 21.16 \\
FSN        & fine-grained video captioning    & clip-sen   & 5 & 1 & 9.39  & 1.67 & 18.30 & -    & -       \\
\textit{Ours} & object-oriented video captioning & object-sen & 73 & 3 & 16.56 & 2.02 & 21.00 & 1.97 & 11.81 \\
\hline
\end{tabular}}
\label{tab:datasets_compare}
\end{table*}

\begin{table}[!t]
\caption{Constituents of most words in our re-annotations.}
\centering
\resizebox{\linewidth}{!}{
\begin{tabular}{ll}
\hline
\multicolumn{1}{l}{} & \multicolumn{1}{c}{Words} \\
\hline
Object&
  \tabincell{l}{boy, girl, man, woman, hopscotch, dog, car, truck, bike, \\motor, bottle, curb, mat, glasses, vest, pants, sweater, ,...} \\ \hline
Activity &
   \tabincell{l}{stand, play, look, jump, run, lean, hug, bend, blow,\\celebrate, clasp, draw, squat, park, lie, lean, wave, ...} \\ \hline
Color               &
  \tabincell{l}{white, red, green, grey, yellow, blue, orange, pink, \\purple, beige, plaid, floral, ...} \\ \hline
Interaction
  & \tabincell{l}{following, towards, back to, right, \\left, in front of, looking, holding, ...}                          \\ \hline
Stuff
 & \tabincell{l}{sidewalk, lawn, concrete ground, playground, cement, \\park, wall, soil, beach, ...} \\ \hline
\end{tabular}}
\label{tab:datasets_show_words}
\end{table}

In this section, we first introduce the dataset collection, followed by the implementation details. Next, our experimental results are reported accompanied by the comparisons with other methods. Finally, the ablation studies are presented to analyze the impact of each component in our OVC-Net.

\subsection{Object-Oriented Video Captioning Dataset}

The most widely used datasets for video captioning are the MSR-Video to Text (MSR-VTT) dataset \cite{20} and the Microsoft Video Description (MSVD) dataset \cite{27}. The MSR-VTT dataset contains 10K short video clips and 200K video-sentence pairs, and the MSVD dataset contains 1970 YouTube clips. The ActivityNet Captions dataset \cite{18}, the most popular benchmark for dense video captioning, provides 20K videos from 200 activity classes (\eg, drinking, dancing, playing games) . We summarize all these datasets in Table \ref{tab:datasets_compare} in terms of type of data, length of sentences, verbs per sentence, adjectives per sentence, \etc \ We can see the type of data in all datasets are clip-sentence pairs, where each video clip has multiple descriptions for different objects. However, object-level specific information is not available, for example, which sentence describes which object. This kind of data goes against learning the functional mappings between vision and language due to the one-to-many nature.

To adapt to the proposed task, and overcome the limitations mentioned above, we re-annotate a portion of videos of the ActivityNet Captions dataset with explicit object-sentence pairs. We re-annotate all videos from the class of ‘playing games’. The videos of this class have more diverse scenes than those of the other classes, and each video contains more individuals and interacting activities. As shown in Table \ref{tab:datasets_compare}, the main difference between the new database and the others is, we include the definite object identities. In our re-annotations, there is one sentence associated with each object the video. Totally, we re-annotate 534 object-sentence pairs, and the average length of each object trajectory is about 248 frames. In our experiments, we utilize 418 clips for training and 116 clips for testing.

In addition, object-oriented video captioning requires more informative descriptions to clearly distinguish different objects. We further conduct detailed statistics comparisons between our captioning annotations and the annotations of other datasets. From Table \ref{tab:datasets_compare}, our sentences contain more words in each sentence, including verbs and adjectives. In MSR-VTT, MSVD and ActivityNet Captions, one sentence has less than 1.4 verbs. The annotated sentences in the Fine-grained Sports Narrative (FSN) Dataset have more verbs to describe fine-grained actions \cite{19}. Nonetheless, even compared with FSN, our sentences provide richer information. Similarly, we analyze the adjectives in the annotated sentences. Each sentence in our re-annotations contains 2 adjectives, however, the sentences from all the other datasets only have less than 0.67 adjectives. The wide difference shows our data are more informative. The ratios of adjectives and verbs is not significantly higher than that of the other datasets, because we have fewer videos than the others, and therefore we can adequately describe the objects using these adjectives. 

\begin{figure}[!t]
\centering
\includegraphics[width=0.85\linewidth]{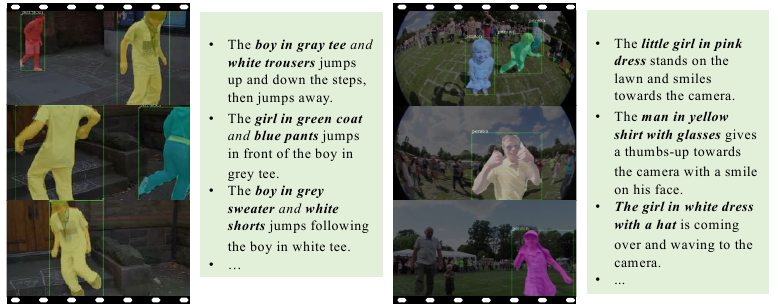}
\caption{Examples of our re-annotations. We label most objects in the video. We will have more object-sentence pairs especially in complex scenarios.}
\label{fig:dataset_example}
\end{figure}

Table \ref{tab:datasets_show_words} shows the constituents of words in our re-annotations. Our sentences contain words which can describe the detailed information for distinguishing easily-confused objects, \eg, color of the clothes (common color, plaid, floral), type of clothes (shirt, sweater, pants, shorts, vest, dress, skirt). Fig. \ref{fig:dataset_example} shows two examples in our database. Our re-annotations will be publicly released upon acceptance.

Although the size of our data is smaller than those of the existing datasets, it is sufficient for this work. So far, we focus on a particular scene of `playing games' to test the learning capabilities and effectiveness of the proposed approach, and its generalization to other scenarios can be investigated in the future.

\subsection{Implementation Details}

\subsubsection{Object trajectory processing}For each object, we sample 40 equally-spaced frames. We adopt VGG19 pretrained on ImageNet as the backbone to extract the visual semantic features from the last pooling layer. While building the object-oriented temporal graph, we feed the cropped objects into the backbone to get their corresponding local features, and feed the frames where the object exists into the backbone to get the global features. Finally, for each channel of RBG, we extract 16-dimensional color histograms, resulting in the final 4144-dim local features and 4096-dim global features.

\subsubsection{Sentence processing}We remove the punctuations, split them with blank space and convert all words into lower-case. We set the maximum length of each sentence to be 25. The sentences longer than 25 are truncated. We randomly initialize the word embedding with a fixed-size of 512.

\subsubsection{Training details}The detail enhancement branch consists of three fully-connected layers. We group all objects which may have activities into three super-classes: human, vehicle and animal. For human, we further split into two sub-classes to improve the distinction: male and female. In the captioning module, the GRU is initialized to have 2 layers with 1024-dimensional hidden units. We empirically set the hyper-parameter $\lambda$ in~\eqref{eq:loss_total} to 0.1. We adopt the adaptive moment estimation (Adam) for optimization. The initialized learning rate is 0.0001.

\subsection{Experimental Results}

\subsubsection{Comparisons with State-of-the-arts}
We adopt the metrics, BLEU (B) \cite{41}, METEOR (M) \cite{44}, ROUGE-L (R) \cite{43}, CIDEr-D (C) \cite{42}, which are widely used in text generation tasks, to quantitatively evaluate the proposed approach. The higher scores represent better performance of the methods. We compare the performance of the proposed OVC-Net with the state-of-the-art methods for traditional video captioning, MP-LSTM \cite{16}, SA-LSTM \cite{15}, S2VT \cite{7}, RecNet \cite{13}, FSTA \cite{6}, and OA-BTG \cite{46}. In traditional video captioning, these methods are trained on video-sentence pairs to generate descriptions for videos. To adapt to the new task, they can remain the general framework, yet directly use the object-sentence pairs to generate the descriptions for objects.

\begin{table}[t]
\centering
\caption{Performance comparisons with state-of-the-art methods for traditional video captioning. \textit{TG} is the object-oriented temporal graph. \textit{DE} is the detail enhancement module.}
\resizebox{\linewidth}{!}{
\begin{tabular}{lccccccc}
\hline
Model      & B@1  & B@2  & B@3  & B@4  & M   & R  & C \\ \hline
MP-LSTM\cite{16} & 43.2 & 29.2 & 21.7 & 16.2 & 18.2 & 40.3  & 38.8   \\
SA-LSTM\cite{15} & 43.7 & 29.8 & 22.0 & 16.3 & 18.3 & 39.7  & 37.3 \\
S2VT\cite{7} & 40.2 & 26.8 & 19.6 & 14.1 & 16.6 & 37.9  & 31.7 \\
RecNet\cite{13} & 42.7 & 28.5 & 21.0 & 15.6 & 17.5 & 40.3  & 42.1 \\
FSTA\cite{6} & 41.6 & 27.1 & 19.2 & 13.5 & 16.8 & 39.0  & 34.3 \\
OA-BTG\cite{46} & 39.4 & 25.7 & 19.2 & 14.5 & 16.1 & 38.6  & 37.9 \\ \hline
\textit{Ours}(TG) & 45.0 & 32.4 & 25.0 & 19.3 & 19.7 & 45.0  & 50.2 \\
\textit{Ours}(TG+DE) & \textbf{47.0} & \textbf{33.4} & \textbf{25.7} & \textbf{20.2} & \textbf{20.0} & \textbf{45.1} & \textbf{50.4} \\ \hline
\end{tabular}}
\label{tab:results}
\end{table}

\begin{table}
\centering
\caption{Ablation study of the components in the object-oriented temporal graph.}
\begin{tabular}{cccc|cccc}
\hline
Global& Local& Color& Spatial&  B@4  & M    & R  & C         \\ \hline
\checkmark      &       &       &          & 16.1          & 16.9          & 39.3          & 38.5          \\
       & \checkmark      &       &         & 16.6          & 18.0          & 42.3          & 45.5          \\
\checkmark       & \checkmark      &       &         & 18.1          & 19.3          & 44.5          & \textbf{52.1}         \\
\checkmark       & \checkmark      & \checkmark      &         &  18.7          & 19.5          & 44.3          & 50.9          \\
\checkmark       & \checkmark      & \checkmark      & \checkmark        &  \textbf{19.3} & \textbf{19.7} & \textbf{45.0} & 50.2\\ \hline
\end{tabular}
\label{tab:ablation_study}
\end{table}

MP-LSTM is a baseline method utilized the mean pooling to process the frames. SA-LSTM introduces the temporal attention to decide the key frames. S2VT adopts LSTM both in the encoder and the decoder. RecNet, FSTA and OA-BTG achieve the state-of-the-art performances of video captioning, however, all these data-driven methods cannot perform well under small-samples for the weak capability in understanding the spatio-temporal transitions as shown in Table \ref{tab:results}.

FSTA adopts the masks generated by Mask R-CNN to filter the objects from the stuff. However, it ignores the stuff and the features with explicit boundary are less conducive to training, hence the performance is not ideal. FSTA adopts Mask R-CNN as the backbone to extract visual semantic features. However, the features extracted via models for object detection pre-trained on MS COCO \cite{64} have less semantic information compared to the features extracted from the models for image classification pre-trained on ImageNet. Therefore, FSTA takes more time to converge. For OA-BTG, owing to the effect of inaccurate trajectories and large amounts of parameters, it cannot obtain a satisfactory performance as well. The MP-LSTM and SA-LSTM perform better than the other methods, because they have fewer parameters and can be trained well under small-samples.

As shown in Table \ref{tab:results}, the BLEU@4 score with building the object-oriented temporal graph only is 19.3, already significantly better than other methods. It proves that the temporal graph can effectively present the objects, including the actions, attributes, interactions with other objects, as well as interactions with the stuff. The construction of the object-oriented temporal graph achieves understanding the activities from the temporal evolution in the visual features and the dynamic movement of the spatial locations, yet not dependence on training under large amounts of data. Therefore, the proposed approach can precisely understand the actions within limited data, and effectively learn the cross-modal relationships.

\begin{figure}[!t]
    \centering
    \includegraphics[width=\linewidth]{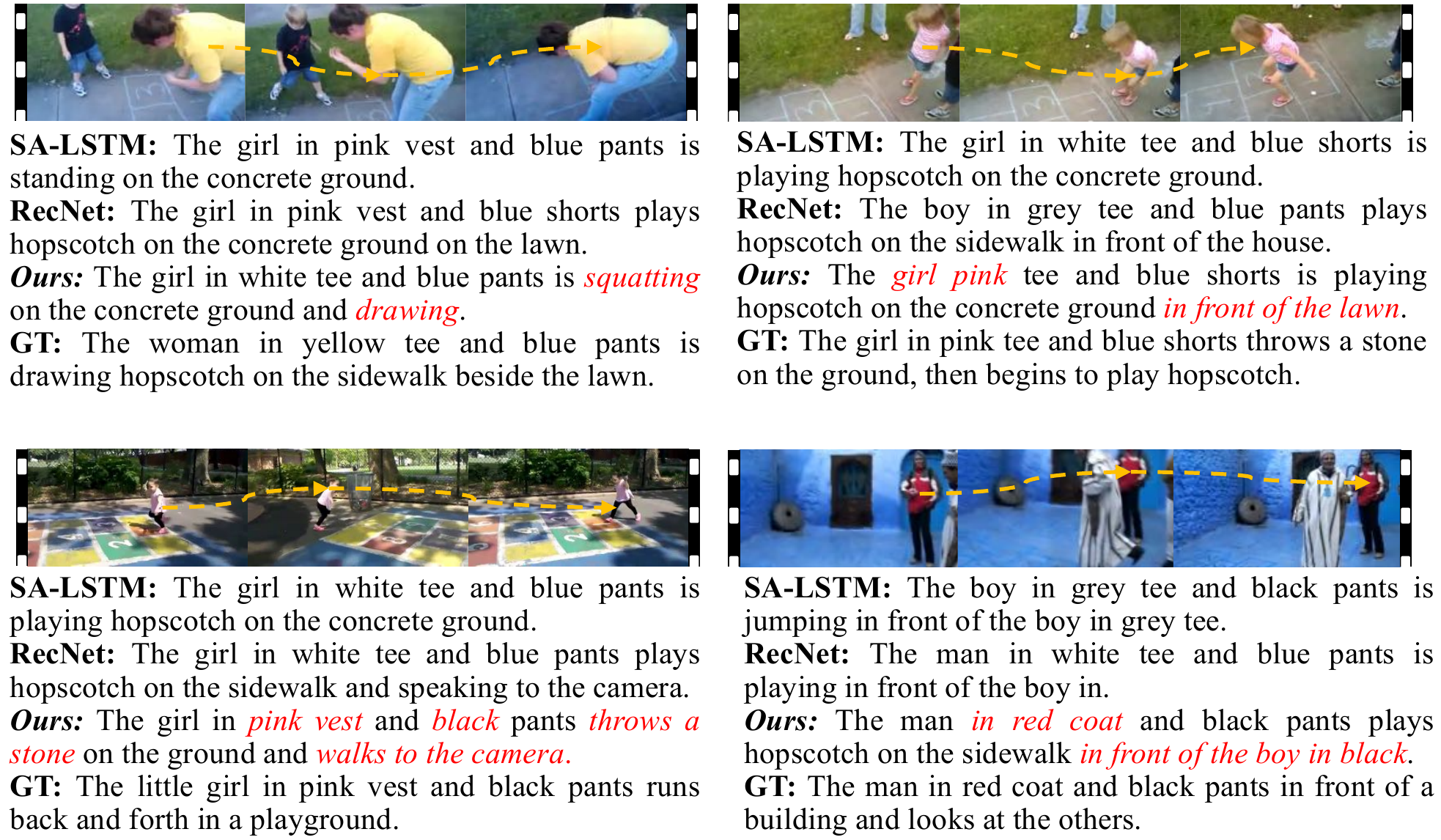}
    \caption{Visualization examples of comparisons with SA-LSTM and RecNet. \textit{GT} is the ground-truth sentence for the object.}
    \label{fig:results}
\end{figure}

After combing the detail enhancement module to revise the detailed descriptions, the BLEU@4 score is further improved to 20.2. From the visualization examples in Fig. \ref{fig:results}, it can be seen that we can generate more accurate descriptions for the actions, like `draw', `throw a stone'. Meanwhile, it can describe the spatial relationships between objects and stuff, and the attributes of objects in details more accurately.

\subsubsection{Ablation Study}

\begin{figure}[!t]
    \centering
    \includegraphics[width=\linewidth]{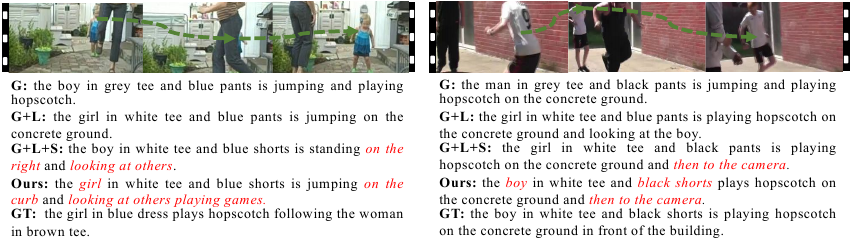}
    \caption{Visualization examples for ablation study. \textit{G}, \textit{L}, \textit{S} denote the global features, the local features, and the spatial location, separately. \textit{Ours} is the proposed full framework.}
    \label{fig:ablation_study_1}
\end{figure}

\begin{figure}[!t]
    \centering
    \includegraphics[width=\linewidth]{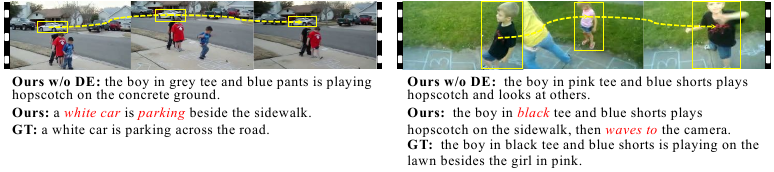}
    \caption{The first is an example of an non-human object. The second example shows we can generate accurate description under an incorrect trajectory.}
    \label{fig:ablation_study_2}
\end{figure}

Shown in Table \ref{tab:results}, while building the object-oriented temporal graph only, we already get better performance than other methods. To further validate the effect of each component in the object-oriented temporal graph, we conduct detailed ablation study in Table \ref{tab:ablation_study}. The BLEU@4 performance with global features is only 16.1, whereas, the performance with local features only is 16.6. It demonstrates that the explicit features of objects are benefit for learning the translational mappings. Adding the color information, the performance is improved to 18.7, which verifies that our color information can effectively solve the problem that the features from top layers of a neural network lack appearance information.
Subsequently, we combine spatial locations to represent the moving trajectories more clearly. From the experimental results, it can be seen that the dynamic movement of spatial locations works great to help understand activities along time.


We show more visualization results for the ablation study in Fig. \ref{fig:ablation_study_1}. While use the global information only, owing to the confusion caused by other concurrent objects, the model cannot effectively bridge the visual contents and natural language, resulting in the muddled generated descriptions. After combining the local features, it can learn the functional relationships better. Adding the spatial locations, the generated descriptions contain more information about the spatial locations, \eg, `in front of', `on the right'. In the second example, the model can reason the boy `back to the camera' from the spatial location evolution. Finally, with the color information and the detail enhancement, the generated attributes become more accurate.

We show an example of describing a non-human object in Fig. \ref{fig:ablation_study_2}. Without the detail enhancement, the model cannot recognize the object is a `car'. After combining the detail enhancement, it can accurately describe the object `car', the status `parking', and the location `across the road'. It proves that the detail enhancement module can effectively improve the model learning. In the second example, when track `the boy in black', ID switches to `the girl in pink' owing to the object occlusion and camera moving. However, we can still accurately describe `the boy in black' under the inaccurate tracking results. Nonetheless, if the tracking results are extremely poor, for example, under heavy occlusion and frequent ID switches, the model may fails to capture accurate visual features, resulting in the overall performance degradation. How to improve the detection and tracking algorithm to adapt to natural videos is an important direction in the future.

Overall, the experimental results indicate that the proposed approach can precisely understand and describe the concurrent activities without temporal cues from other tasks. The object-oriented temporal graph is effective for presenting the actions of the objects, the object-object interactions, and the object-stuff interactions. Meanwhile, the detail enhancement can significantly facilitate the model training and revise the generated detailed descriptions. Moreover, the explicit object-sentence pairs benefit for learning the functional mappings. So far, we focus on the scene of `playing games', however, we believe our framework can be extended to other scenarios.

\section{Conclusions}

In this paper, we propose a novel task of object-oriented video captioning, which transforms video-level video captioning to object-level analyses. Unlike previous image-based data-driven methods, the proposed video-based framework aims at more realistic video understanding for the purpose of analyzing the concurrent activities. We designed an object-oriented video captioning network (OVC)-Net via temporal graph and detail enhancement. The proposed network can analyze the actions, the object-object interactions, as well as the object-stuff interactions, from the spatio-temporal evolutions in the visual features and spatial locations. To the best of our knowledge, this is the first work extends the video captioning from generating a holistic description to detailed descriptions for specific objects. The re-annotated database is the first dataset that provides the definite object-sentence pairs, with which further works can explore the vision-language relationships better. Moreover, the proposed network is the first video-based framework for video captioning, attaining real video understanding. Experimental results demonstrate that the proposed method achieve significant improvement than the state-of-the-art methods for traditional video captioning. It is also indicated that the proposed framework can provide richer information of the whole scene and achieve stable performance under small-samples.


\bibliographystyle{IEEEtran}
\bibliography{IEEEabrv,main}

\begin{thebibliography}{10}
\providecommand{\url}[1]{#1}
\csname url@samestyle\endcsname
\providecommand{\newblock}{\relax}
\providecommand{\bibinfo}[2]{#2}
\providecommand{\BIBentrySTDinterwordspacing}{\spaceskip=0pt\relax}
\providecommand{\BIBentryALTinterwordstretchfactor}{4}
\providecommand{\BIBentryALTinterwordspacing}{\spaceskip=\fontdimen2\font plus
\BIBentryALTinterwordstretchfactor\fontdimen3\font minus
  \fontdimen4\font\relax}
\providecommand{\BIBforeignlanguage}[2]{{%
\expandafter\ifx\csname l@#1\endcsname\relax
\typeout{** WARNING: IEEEtran.bst: No hyphenation pattern has been}%
\typeout{** loaded for the language `#1'. Using the pattern for}%
\typeout{** the default language instead.}%
\else
\language=\csname l@#1\endcsname
\fi
#2}}
\providecommand{\BIBdecl}{\relax}
\BIBdecl

\bibitem{1}
H.~Yu, J.~Wang, Z.~Huang, Y.~Yang, and W.~Xu, ``Video paragraph captioning
  using hierarchical recurrent neural networks,'' in \emph{Proceedings of the
  IEEE conference on computer vision and pattern recognition}, 2016, pp.
  4584--4593.

\bibitem{2}
Q.~Wu, C.~Shen, L.~Liu, A.~Dick, and A.~Van Den~Hengel, ``What value do
  explicit high level concepts have in vision to language problems?'' in
  \emph{Proceedings of the IEEE conference on computer vision and pattern
  recognition}, 2016, pp. 203--212.

\bibitem{3}
M.~Luo, X.~Chang, Z.~Li, L.~Nie, A.~G. Hauptmann, and Q.~Zheng, ``Simple to
  complex cross-modal learning to rank,'' \emph{Computer Vision and Image
  Understanding}, vol. 163, pp. 67--77, 2017.

\bibitem{4}
Z.~Zeng, Z.~Li, D.~Cheng, H.~Zhang, K.~Zhan, and Y.~Yang, ``Two-stream
  multirate recurrent neural network for video-based pedestrian
  reidentification,'' \emph{IEEE Transactions on Industrial Informatics},
  vol.~14, no.~7, pp. 3179--3186, 2017.

\bibitem{5}
Z.~Li, F.~Nie, X.~Chang, L.~Nie, H.~Zhang, and Y.~Yang, ``Rank-constrained
  spectral clustering with flexible embedding,'' \emph{IEEE transactions on
  neural networks and learning systems}, vol.~29, no.~12, pp. 6073--6082, 2018.

\bibitem{6}
A.-A. Liu, Y.~Qiu, Y.~Wong, Y.-T. Su, and M.~Kankanhalli, ``A fine-grained
  spatial-temporal attention model for video captioning,'' \emph{IEEE Access},
  vol.~6, pp. 68\,463--68\,471, 2018.

\bibitem{7}
B.~Wang, L.~Ma, W.~Zhang, and W.~Liu, ``Reconstruction network for video
  captioning,'' in \emph{Proceedings of the IEEE Conference on Computer Vision
  and Pattern Recognition}, 2018, pp. 7622--7631.

\bibitem{11}
C.~Hori, T.~Hori, T.-Y. Lee, Z.~Zhang, B.~Harsham, J.~R. Hershey, T.~K. Marks,
  and K.~Sumi, ``Attention-based multimodal fusion for video description,'' in
  \emph{Proceedings of the IEEE international conference on computer vision},
  2017, pp. 4193--4202.

\bibitem{13}
S.~Venugopalan, M.~Rohrbach, J.~Donahue, R.~Mooney, T.~Darrell, and K.~Saenko,
  ``Sequence to sequence-video to text,'' in \emph{Proceedings of the IEEE
  international conference on computer vision}, 2015, pp. 4534--4542.

\bibitem{15}
L.~Yao, A.~Torabi, K.~Cho, N.~Ballas, C.~Pal, H.~Larochelle, and A.~Courville,
  ``Describing videos by exploiting temporal structure,'' in \emph{Proceedings
  of the IEEE international conference on computer vision}, 2015, pp.
  4507--4515.

\bibitem{16}
S.~Venugopalan, H.~Xu, J.~Donahue, M.~Rohrbach, R.~Mooney, and K.~Saenko,
  ``Translating videos to natural language using deep recurrent neural
  networks,'' \emph{arXiv preprint arXiv:1412.4729}, 2014.

\bibitem{17}
J.~Donahue, L.~Anne~Hendricks, S.~Guadarrama, M.~Rohrbach, S.~Venugopalan,
  K.~Saenko, and T.~Darrell, ``Long-term recurrent convolutional networks for
  visual recognition and description,'' in \emph{Proceedings of the IEEE
  conference on computer vision and pattern recognition}, 2015, pp. 2625--2634.

\bibitem{18}
R.~Krishna, K.~Hata, F.~Ren, L.~Fei-Fei, and J.~Carlos~Niebles,
  ``Dense-captioning events in videos,'' in \emph{Proceedings of the IEEE
  international conference on computer vision}, 2017, pp. 706--715.

\bibitem{20}
J.~Xu, T.~Mei, T.~Yao, and Y.~Rui, ``Msr-vtt: A large video description dataset
  for bridging video and language,'' in \emph{Proceedings of the IEEE
  conference on computer vision and pattern recognition}, 2016, pp. 5288--5296.

\bibitem{46}
J.~Zhang and Y.~Peng, ``Object-aware aggregation with bidirectional temporal
  graph for video captioning,'' in \emph{Proceedings of the IEEE Conference on
  Computer Vision and Pattern Recognition}, 2019, pp. 8327--8336.

\bibitem{31}
D.~Tran, L.~Bourdev, R.~Fergus, L.~Torresani, and M.~Paluri, ``Learning
  spatiotemporal features with 3d convolutional networks,'' in
  \emph{Proceedings of the IEEE international conference on computer vision},
  2015, pp. 4489--4497.

\bibitem{32}
K.~Hara, H.~Kataoka, and Y.~Satoh, ``Can spatiotemporal 3d cnns retrace the
  history of 2d cnns and imagenet?'' in \emph{Proceedings of the IEEE
  conference on Computer Vision and Pattern Recognition}, 2018, pp. 6546--6555.

\bibitem{33}
------, ``Learning spatio-temporal features with 3d residual networks for
  action recognition,'' in \emph{Proceedings of the IEEE International
  Conference on Computer Vision}, 2017, pp. 3154--3160.

\bibitem{19}
H.~Yu, S.~Cheng, B.~Ni, M.~Wang, J.~Zhang, and X.~Yang, ``Fine-grained video
  captioning for sports narrative,'' in \emph{Proceedings of the IEEE
  Conference on Computer Vision and Pattern Recognition}, 2018, pp. 6006--6015.

\bibitem{27}
D.~L. Chen and W.~B. Dolan, ``Collecting highly parallel data for paraphrase
  evaluation,'' in \emph{Proceedings of the 49th Annual Meeting of the
  Association for Computational Linguistics: Human Language Technologies-Volume
  1}.\hskip 1em plus 0.5em minus 0.4em\relax Association for Computational
  Linguistics, 2011, pp. 190--200.

\bibitem{41}
K.~Papineni, S.~Roukos, T.~Ward, and W.-J. Zhu, ``Bleu: a method for automatic
  evaluation of machine translation,'' in \emph{Proceedings of the 40th annual
  meeting on association for computational linguistics}.\hskip 1em plus 0.5em
  minus 0.4em\relax Association for Computational Linguistics, 2002, pp.
  311--318.

\bibitem{42}
R.~Vedantam, C.~Lawrence~Zitnick, and D.~Parikh, ``Cider: Consensus-based image
  description evaluation,'' in \emph{Proceedings of the IEEE conference on
  computer vision and pattern recognition}, 2015, pp. 4566--4575.

\bibitem{43}
C.-Y. Lin, ``Rouge: A package for automatic evaluation of summaries,'' in
  \emph{Text summarization branches out}, 2004, pp. 74--81.

\bibitem{44}
S.~Banerjee and A.~Lavie, ``Meteor: An automatic metric for mt evaluation with
  improved correlation with human judgments,'' in \emph{Proceedings of the acl
  workshop on intrinsic and extrinsic evaluation measures for machine
  translation and/or summarization}, 2005, pp. 65--72.

\bibitem{8}
K.~Xu, J.~Ba, R.~Kiros, K.~Cho, A.~Courville, R.~Salakhudinov, R.~Zemel, and
  Y.~Bengio, ``Show, attend and tell: Neural image caption generation with
  visual attention,'' in \emph{International conference on machine learning},
  2015, pp. 2048--2057.

\bibitem{9}
P.~Anderson, X.~He, C.~Buehler, D.~Teney, M.~Johnson, S.~Gould, and L.~Zhang,
  ``Bottom-up and top-down attention for image captioning and visual question
  answering,'' in \emph{Proceedings of the IEEE Conference on Computer Vision
  and Pattern Recognition}, 2018, pp. 6077--6086.

\bibitem{10}
L.~Chen, H.~Zhang, J.~Xiao, L.~Nie, J.~Shao, W.~Liu, and T.-S. Chua, ``Sca-cnn:
  Spatial and channel-wise attention in convolutional networks for image
  captioning,'' in \emph{Proceedings of the IEEE conference on computer vision
  and pattern recognition}, 2017, pp. 5659--5667.

\bibitem{14}
C.~Liu, J.~Mao, F.~Sha, and A.~Yuille, ``Attention correctness in neural image
  captioning,'' in \emph{Thirty-First AAAI Conference on Artificial
  Intelligence}, 2017.

\bibitem{12}
J.~Johnson, A.~Karpathy, and L.~Fei-Fei, ``Densecap: Fully convolutional
  localization networks for dense captioning,'' in \emph{Proceedings of the
  IEEE Conference on Computer Vision and Pattern Recognition}, 2016, pp.
  4565--4574.

\bibitem{36}
S.~Guadarrama, N.~Krishnamoorthy, G.~Malkarnenkar, S.~Venugopalan, R.~Mooney,
  T.~Darrell, and K.~Saenko, ``Youtube2text: Recognizing and describing
  arbitrary activities using semantic hierarchies and zero-shot recognition,''
  in \emph{Proceedings of the IEEE international conference on computer
  vision}, 2013, pp. 2712--2719.

\bibitem{37}
M.~Rohrbach, W.~Qiu, I.~Titov, S.~Thater, M.~Pinkal, and B.~Schiele,
  ``Translating video content to natural language descriptions,'' in
  \emph{Proceedings of the IEEE International Conference on Computer Vision},
  2013, pp. 433--440.

\bibitem{38}
J.~Song, Z.~Guo, L.~Gao, W.~Liu, D.~Zhang, and H.~T. Shen, ``Hierarchical lstm
  with adjusted temporal attention for video captioning,'' \emph{arXiv preprint
  arXiv:1706.01231}, 2017.

\bibitem{39}
S.~Hochreiter and J.~Schmidhuber, ``Long short-term memory,'' \emph{Neural
  computation}, vol.~9, no.~8, pp. 1735--1780, 1997.

\bibitem{60}
M.~Frid-Adar, E.~Klang, M.~Amitai, J.~Goldberger, and H.~Greenspan, ``Synthetic
  data augmentation using gan for improved liver lesion classification,'' in
  \emph{2018 IEEE 15th international symposium on biomedical imaging (ISBI
  2018)}.\hskip 1em plus 0.5em minus 0.4em\relax IEEE, 2018, pp. 289--293.

\bibitem{61}
Y.~Yang, J.~Zhou, J.~Ai, Y.~Bin, A.~Hanjalic, H.~T. Shen, and Y.~Ji, ``Video
  captioning by adversarial lstm,'' \emph{IEEE Transactions on Image
  Processing}, vol.~27, pp. 5600--5611, 2018.

\bibitem{62}
N.~Aafaq, A.~Mian, W.~Liu, S.~Z. Gilani, and M.~Shah, ``Video description: A
  survey of methods, datasets, and evaluation metrics,'' \emph{ACM Computing
  Surveys (CSUR)}, vol.~52, no.~6, pp. 1--37, 2019.

\bibitem{28}
J.~Deng, W.~Dong, R.~Socher, L.-J. Li, K.~Li, and L.~Fei-Fei, ``Imagenet: A
  large-scale hierarchical image database,'' in \emph{2009 IEEE conference on
  computer vision and pattern recognition}.\hskip 1em plus 0.5em minus
  0.4em\relax Ieee, 2009, pp. 248--255.

\bibitem{21}
J.~Redmon and A.~Farhadi, ``Yolov3: An incremental improvement,'' \emph{arXiv
  preprint arXiv:1804.02767}, 2018.

\bibitem{22}
S.~Ren, K.~He, R.~Girshick, and J.~Sun, ``Faster r-cnn: Towards real-time
  object detection with region proposal networks,'' in \emph{Advances in neural
  information processing systems}, 2015, pp. 91--99.

\bibitem{25}
K.~He, G.~Gkioxari, P.~Doll{\'a}r, and R.~Girshick, ``Mask r-cnn,'' in
  \emph{Proceedings of the IEEE international conference on computer vision},
  2017, pp. 2961--2969.

\bibitem{47}
P.~Bergmann, T.~Meinhardt, and L.~Leal-Taixe, ``Tracking without bells and
  whistles,'' \emph{arXiv preprint arXiv:1903.05625}, 2019.

\bibitem{24}
G.~Wang, Y.~Wang, H.~Zhang, R.~Gu, and J.-N. Hwang, ``Exploit the connectivity:
  Multi-object tracking with trackletnet,'' in \emph{Proceedings of the 27th
  ACM International Conference on Multimedia}.\hskip 1em plus 0.5em minus
  0.4em\relax ACM, 2019, pp. 482--490.

\bibitem{48}
\BIBentryALTinterwordspacing
L.~Yang, Y.~Fan, and N.~Xu, ``Video instance segmentation,'' \emph{CoRR}, vol.
  abs/1905.04804, 2019. [Online]. Available:
  \url{https://arxiv.org/abs/1905.04804}
\BIBentrySTDinterwordspacing

\bibitem{40}
K.~Cho, B.~Van~Merri{\"e}nboer, C.~Gulcehre, D.~Bahdanau, F.~Bougares,
  H.~Schwenk, and Y.~Bengio, ``Learning phrase representations using rnn
  encoder-decoder for statistical machine translation,'' \emph{arXiv preprint
  arXiv:1406.1078}, 2014.

\bibitem{64}
T.-Y. Lin, M.~Maire, S.~Belongie, J.~Hays, P.~Perona, D.~Ramanan,
  P.~Doll{\'a}r, and C.~L. Zitnick, ``Microsoft coco: Common objects in
  context,'' in \emph{Computer Vision -- ECCV 2014}, D.~Fleet, T.~Pajdla,
  B.~Schiele, and T.~Tuytelaars, Eds.\hskip 1em plus 0.5em minus 0.4em\relax
  Cham: Springer International Publishing, 2014, pp. 740--755.

\end{thebibliography}

\end{document}